\documentclass{ieee}
\usepackage{times}
\usepackage{graphicx}   
\usepackage{colortbl}
\definecolor{cLightRed}{rgb}{1,.70,.70}
\definecolor{cLightYellow}{rgb}{.90,.85,.55}
\definecolor{cLightGray}{rgb}{.90,.90,.90}
\definecolor{cMediumGray}{rgb}{.70,.70,.70}
\definecolor{cDarkGray}{rgb}{.50,.60,.70}
\usepackage{array}
\usepackage{multirow}
\usepackage{tabularx}
\usepackage{amsmath}
\usepackage{amssymb}
\usepackage{subfigure}

\usepackage{algorithmic}

\begin{document}

\title{A New Algorithm for Interactive Structural Image Segmentation}

\author{Alexandre Noma$^1$, Ana B. V. Graciano$^1$, Luis Augusto Consularo$^2$, Roberto M. Cesar-Jr$^1$, Isabelle Bloch$^3$\\
IME-USP\\
$^1$ Department of Computer Science - IME - University of S\~{a}o Paulo - S\~{a}o Paulo, Brazil - \\
{\tt \small \{noma, cesar, abvg\}@ime.usp.br}\\
$^2$ UNIMEP - Methodist University of Piracicaba - Piracicaba-SP, Brazil - {\tt \small laconsul@unimep.br}\\
$^3$ Ecole Nationale Sup\'{e}rieure des T\'{e}l\'{e}communications - CNRS UMR 5141 - Paris, France - \\
{\tt \small Isabelle.Bloch@enst.fr}
}

\maketitle

\begin{abstract}
This paper proposes a novel algorithm for the problem of structural image segmentation through an interactive model-based approach. Interaction is expressed in the model creation, which is done according to user traces drawn over a given input image. Both model and input are then represented by means of attributed relational graphs derived on the fly. Appearance features are taken into account as object attributes and structural properties are expressed as relational attributes. To cope with possible topological differences between both graphs, a new structure called the \textit{deformation} graph is introduced. The segmentation process corresponds to finding a labelling of the input graph that minimizes the deformations introduced in the model when it is updated with input information. This approach has shown to be faster than other segmentation methods, with competitive output quality. Therefore, the method solves the problem of multiple label segmentation in an efficient way. Encouraging results on both natural and target-specific color images, as well as examples showing the reusability of the model, are presented and discussed.
\end{abstract}

\section{Introduction}

Semi-automated, or interactive, image segmentation methods have successfully been used in different applications, whenever human knowledge may be provided as initial guiding clues for the segmentation process. Examples of such methods are the region-growing technique, marker-based watersheds~\cite{soille99}, the IFT~\cite{FalcaoTPAMI04}, graph-cuts and Markov-random fields~\cite{BlakeRBPT04, Rother04, sinop2007seeded}, amongst others. 

Another source of \textit{a priori} information for segmentation are image models, which consist of representative instances of desired objects, conveying different types of features (e.g. color, shape, geometry, relations, etc.) that describe such entities. Approaches guided by models are widely used for a variety of image processing purposes such as medical imaging~\cite{freedman, niessen, olabarriaga01, perchant99bis}, face recognition and tracking~\cite{cesarpatrec05, decarlo02}, and OCR~\cite{lee03}.  

Though the aforementioned interactive approaches have established remarkable contributions to the image segmentation domain, most of them have not attempted to consider image structure to aid the segmentation procedure. An \textit{attributed relational graphs} (ARG) is a particularly useful representation not only for embedding structural information when modelling a problem, but also for expressing appearance features. 

\begin{figure}[t]
\begin{center}
\begin{tabular}{cc}
   \includegraphics[width=4cm]{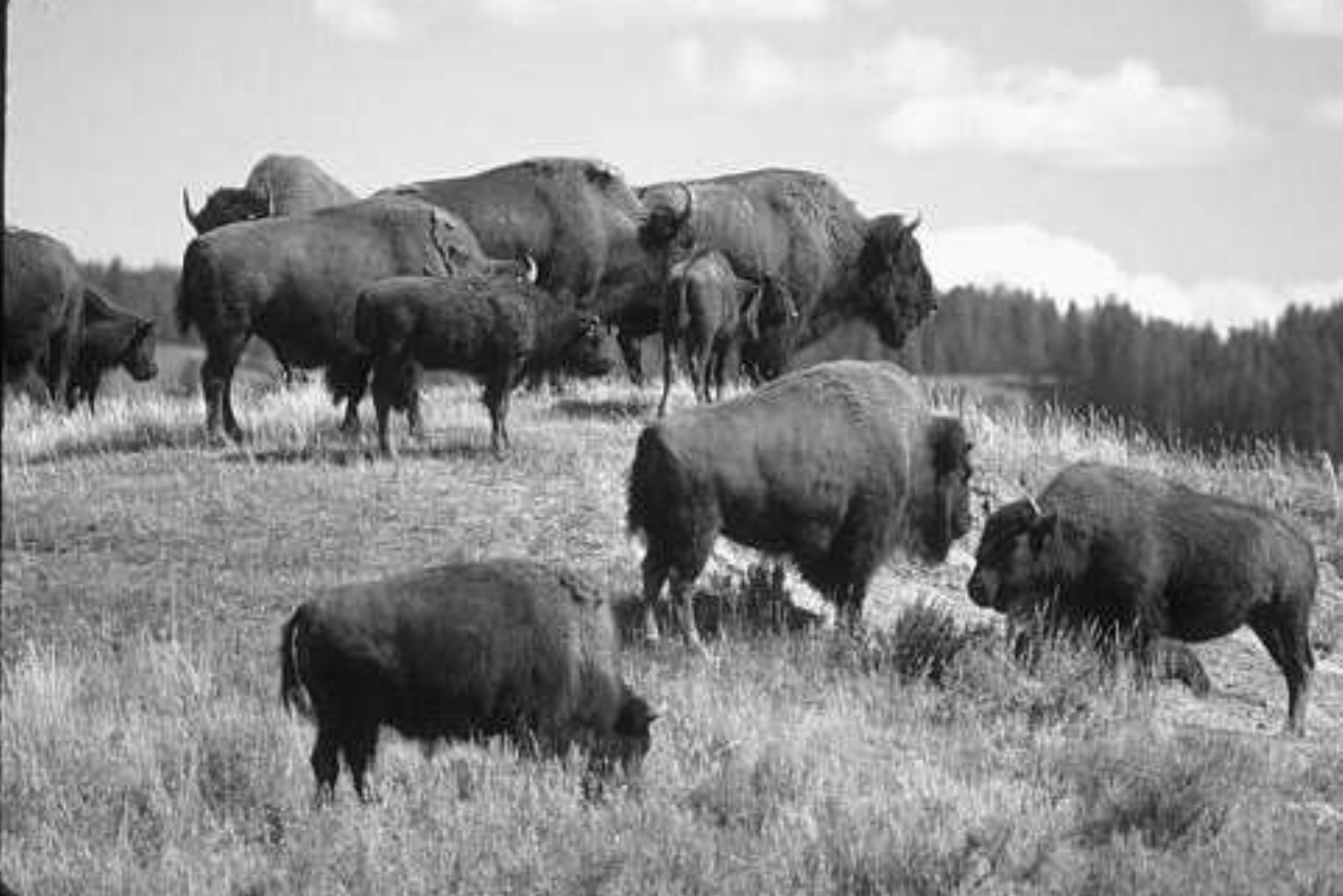} 
   \includegraphics[width=4cm]{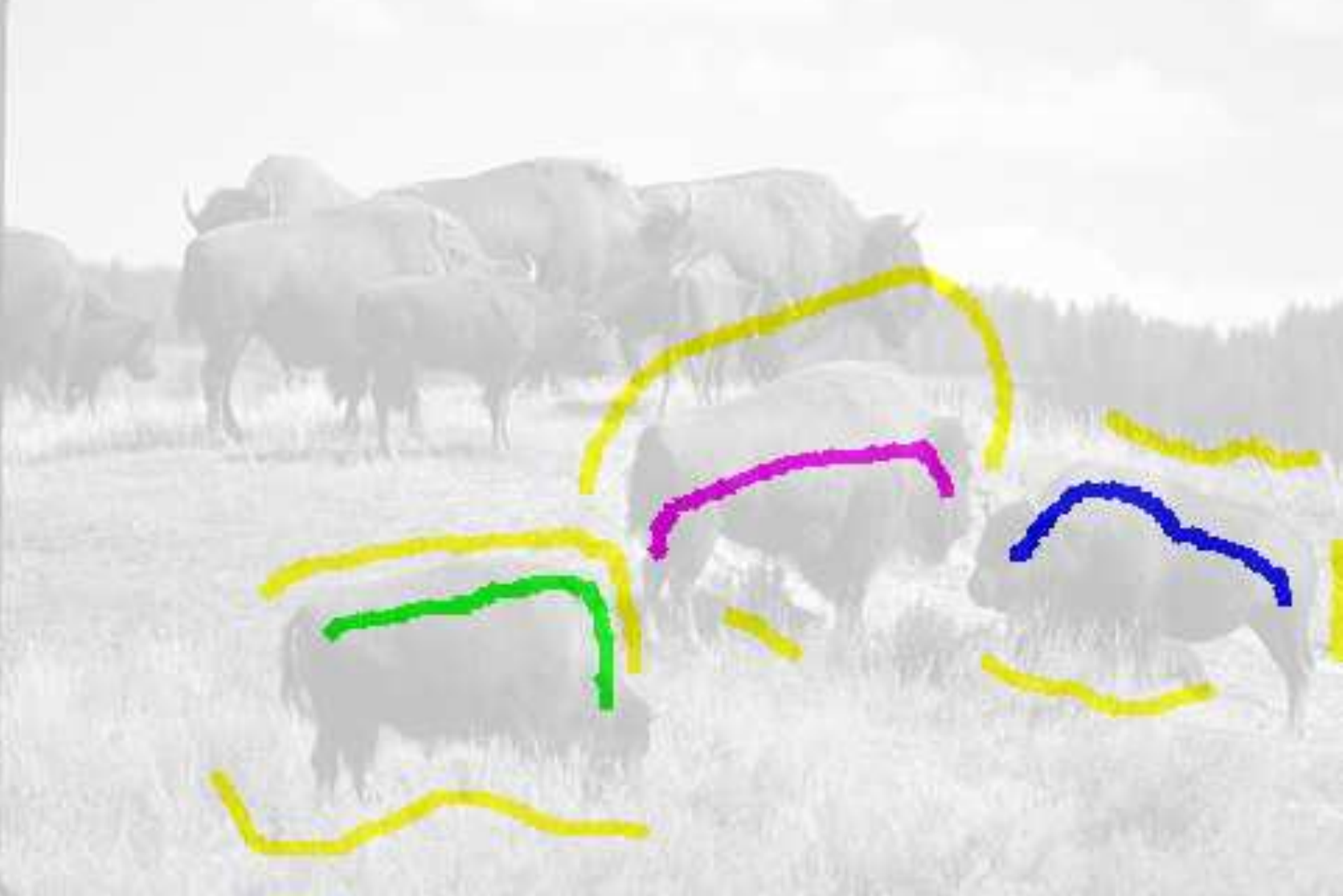} \\
   \includegraphics[width=4cm]{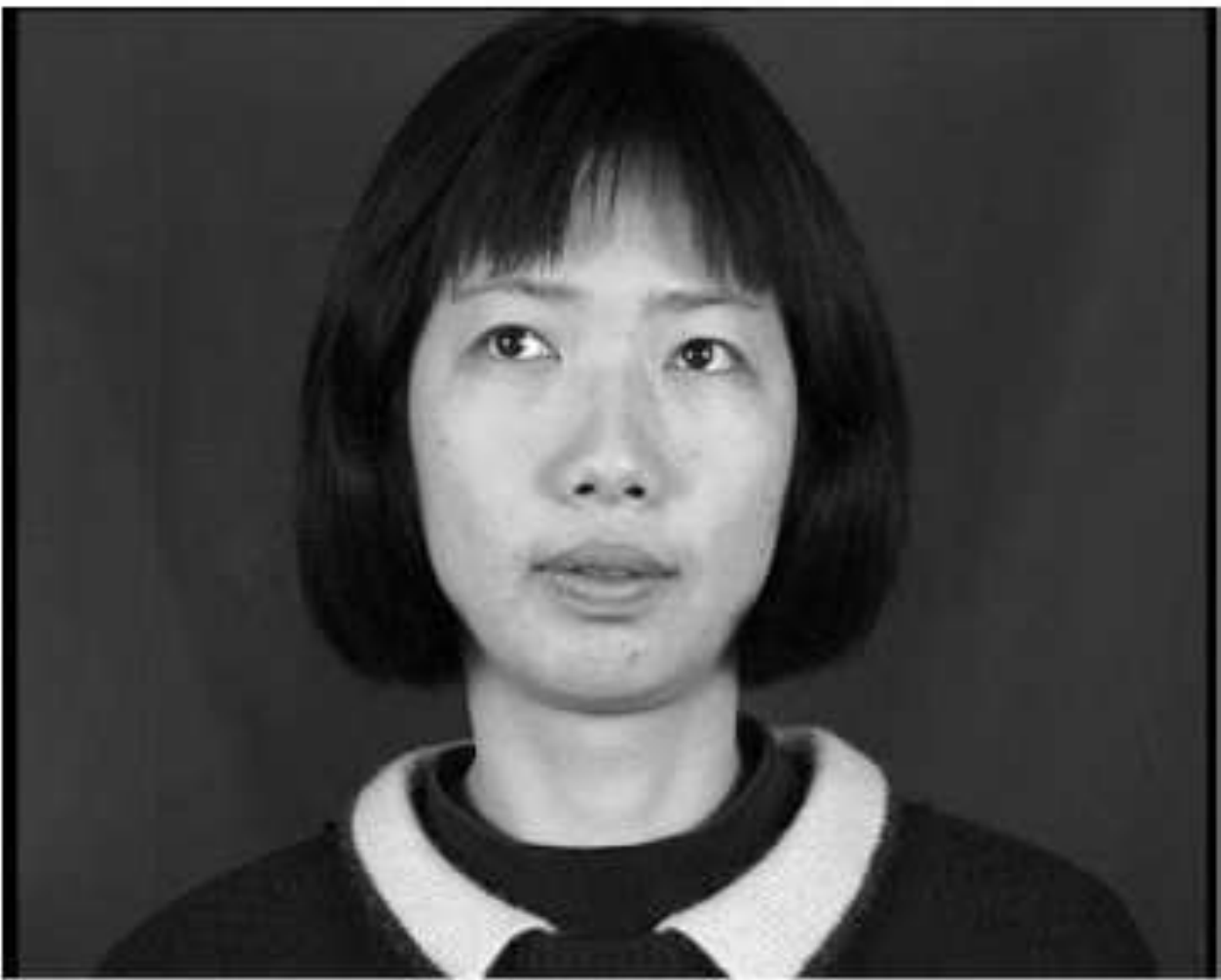} 
   \includegraphics[width=4cm]{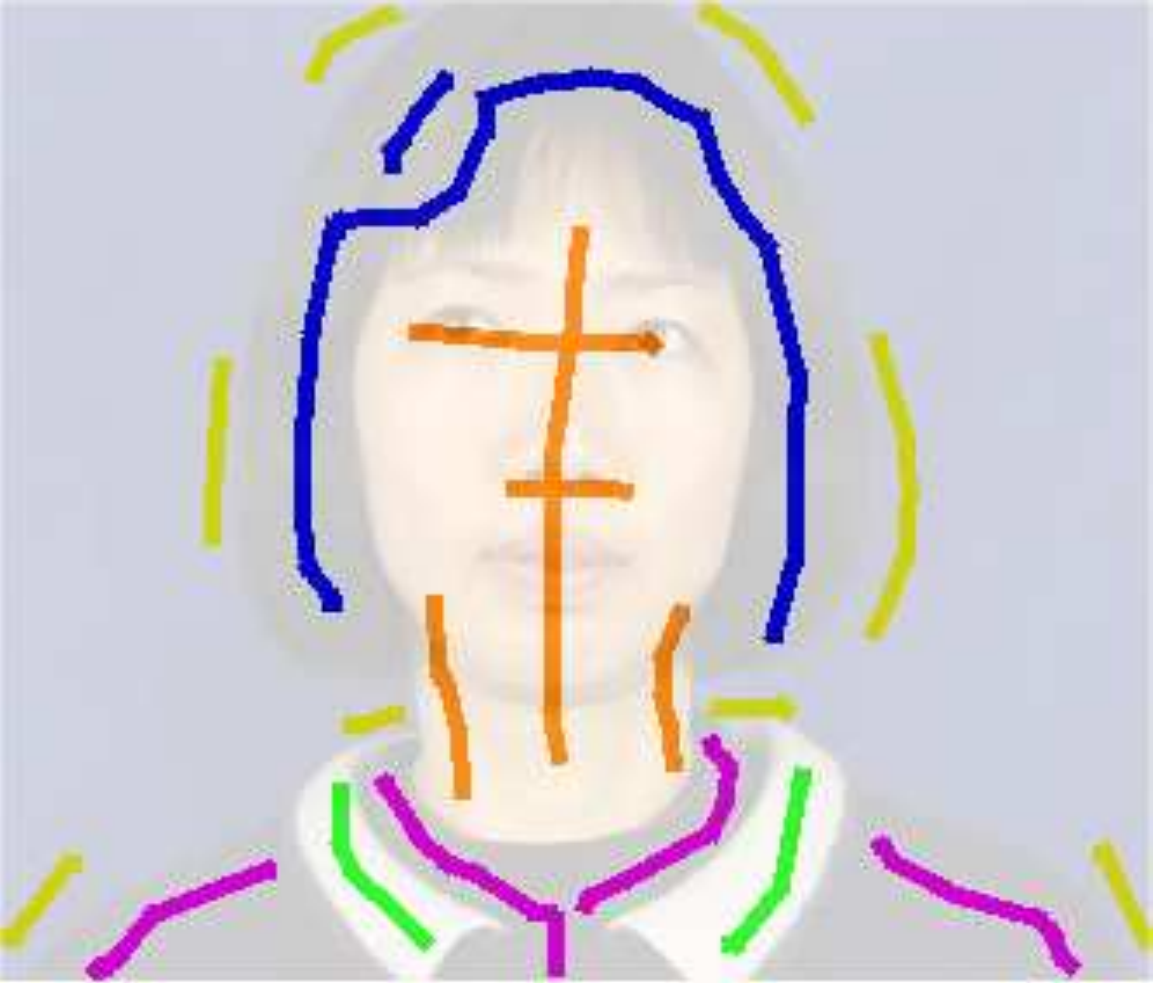} \\
\end{tabular}
\end{center}
\caption{\label{fig:objects} Objects as parts: in this definition, an object might be a whole scene subdivided into parts (buffalos, background), or a single entity (a person) subdivided into meaningful parts (face, clothing, hair). All parts are defined by the user through traces over the regions of interest, and each color represents an object label.}
\end{figure}

\indent Regarding the segmentation issue, the present paper proposes a new algorithm for segmenting color images using both interactive cues and a model using an ARG-based representation. An object (fig.~\ref{fig:objects} left) is considered to be a set of parts (subset of pixels of an image) and their relations. An object model image is defined by a user according to traces drawn over the input (fig.~\ref{fig:objects} right). The input and model images are then represented by means of attributed relational graphs, in which objects and their relations are represented, respectively, as vertices and edges. Under this formulation, the segmentation problem is viewed as a graph matching procedure. 

\begin{figure}[t]
\begin{center}
   \includegraphics[width=8cm]{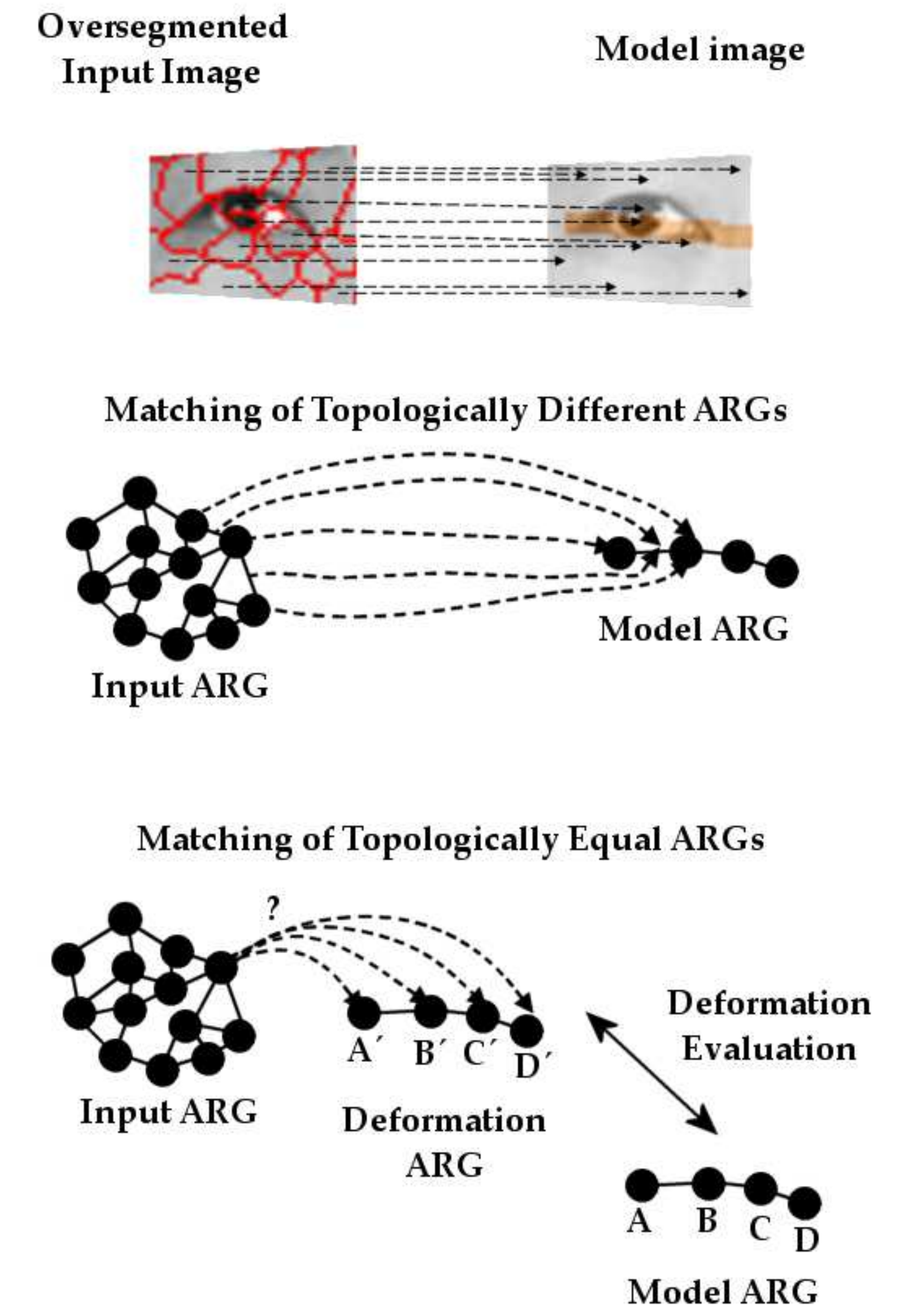} 
\end{center}
\caption{\label{fig:topo} Matching of two topologically different attributed relational graphs versus two topologically equal ones under a deformation point of view.}
\end{figure}

The introduced algorithm substantially improves the approach described in~\cite{consulcesarblochICIP07}, in which structure was taken into account when segmenting an input gray-scale image according to a model, but the structures under comparison, represented by ARGs, often presented different topologies. Such differences imply difficulties to determine a suitable mapping between the input and model graphs, as well as high computational cost. As fig.~\ref{fig:topo} shows, the graph matching problem allows many possible solutions and therefore the optimization procedure not only has to consider vertex similarities, but it also has to evaluate various structural match configurations in order to rule out those which are not plausible for a final segmentation. However, this might be misleading when both topologies are distinct and cause the method to look down on potential solutions, such as when evaluating a match between an input edge connecting two input vertices which represent adjacent oversegmented regions related to two distinct objects and a model edge connecting model vertices which represent the the same previous objects.



In this paper, we propose a novel algorithm for the graph matching step. Each possible matching from an input ARG vertex to a model ARG vertex is seen as a deformation of the model graph (fig.~\ref{fig:topo}), expressed by the introduction of the \textit{deformation} ARG, which represents an altered version of the model ARG that preserves its topological properties while entailing attributes from the given input vertex. This new interpretation addresses the problem of matching two topologically different structures and results in a significantly faster segmentation method.






\indent This paper is organized as follows. Section~\ref{sec:graphdef} presents our formal definition of attributed relational graphs for image representation. Section~\ref{sec:overview} gives an overview of all the steps of the segmentation method proposed herein. Section~\ref{sec:theory} describes the problem of image segmentation as a graph matching task, whereas Section~\ref{sec:algo} introduces the proposed segmentation algorithm based on an optimization technique for matching the input and model graphs. Finally, experimental results are discussed in Section~\ref{sec:results} and a few conclusions, as well as suggestions for future work, are the topic of Section~\ref{sec:conclusion}.

\section{Graph-based representation} 
\label{sec:graphdef}

The representation of images using graphs and the matter of graph matching applied to pattern recognition and image processing have been explored in a variety of situations, such as those reported in the works of Bunke~\cite{Bunke00} and Conte et al~\cite{ConteFSV04}, as well as in the method proposed by Felzenzswalb and Huttenlocher~\cite{felz2}, and the classic work of Wilson and Hancock~\cite{hancock97}, among others.

In this paper, an \textit{attributed relational graph} (ARG) is a directed graph formally expressed as a tuple $G = (V, E, \mu, \nu)$, where $V$ stands for its set of vertices and $E \; \subseteq \; V \times V$, its set of edges. Typically, a vertex represents a single image region (subset of image pixels) and an edge is created between vertices representing two image regions. $\mu : V \rightarrow L_V$ assigns an \textit{object} attribute vector to each vertex of\ $V$, whereas $\nu : E \rightarrow L_E$ assigns a \textit{relational} attribute vector to each edge of $E$.$|V|$ denotes the number of vertices in $V$,
while $|E|$ denotes the number of edges in $E$.

For color images, the object attribute vector is composed of the three average RGB values which characterize the corresponding image region, i.e. $\mu(v) = (R_v, G_v, B_v)$. When dealing with gray-scale images, $\mu(v) = (g(v))$, where $g(v)$ denotes the average gray-level of the
image region associated to vertex $v \in V$. Each component of $\mu(v)$ is normalized between 0 and 1 with respect to the minimum and maximum possible
gray-levels. Similarly, the relational attribute of an edge $e = (v, w) \in E$, $v, w \in V$, is defined as $\nu(e) = (p_w - p_v) / (2d_{max})$, where $p_v$ and $p_w$ are the centroids of their respective corresponding image regions. The factor $d_{max}$ is the largest distance between any two points of the input image region. Other attributes may easily be employed, since the methodology presented herein does not impose any restriction on the nature of $\mu$ and $\nu$.

For the purpose of the segmentation method, three instances of such ARGs are considered: an \textit{input} ARG $G_i = (V_i, E_i, \mu_i, \nu_i)$, derived from the input image, a \textit{model} ARG $G_m = (V_m, E_m, \mu_m, \nu_m)$, representing the objects of interest selected by the user, and a \textit{deformation} ARG $G_d = (V_d, E_d, \mu_d, \nu_d)$, used as an auxiliary data structure for measuring deformations implied in the model when matching a vertex $v \in V_i$ to another $w \in V_m$. 

Subscripts shall be used to denote the corresponding graph, e.g.  $v_i \in V_i$ denotes a
vertex of $G_i$, whereas $(v_i,w_i) \in E_i$ denotes an edge of $G_i$.
Similar notations are used for $G_m$ and $G_d$ as well. 

\section{Methodology overview}
\label{sec:overview}

\begin{figure}[t]
\begin{center}
   \includegraphics[width=8cm]{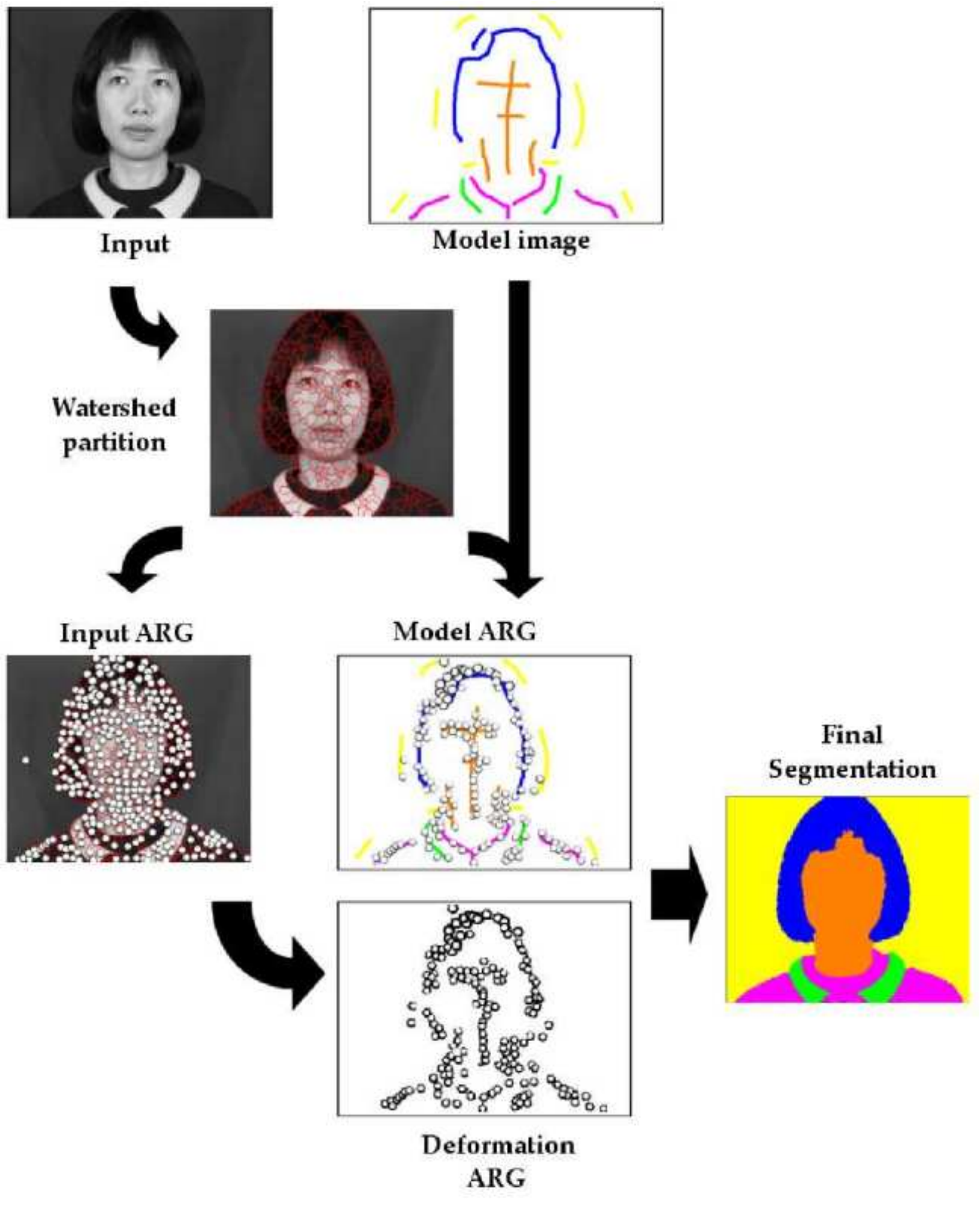} 
\end{center}

\caption{\label{fig:flow} Overview of the methodology steps. The ARGs are depicted only with their vertices for better visualization.}
\end{figure}

The segmentation process is depicted step-by-step in fig.~\ref{fig:flow}. Given an input image to be segmented, the user first points out the target objects by placing traces over the input, thus creating a model image in which each color identifies an object of interest. Next, an oversegmentation is performed using the watershed algorithm to obtain a partition image where the real contours of each object are present.

This oversegmented image is used to create both an input ARG $G_{i}$ and a model ARG $G_m$. The first is obtained in the following fashion: each watershed region gives rise to a vertex and its attributes, whereas adjacent regions devise an edge and its respective attributes. $G_m$ is obtained similarly, but only those watershed regions which intercept the user-defined traces result in a model vertex. Clearly, the input and model ARGs present different topologies and this fact must be accounted for when using structure as a segmentation guide. 

Since the topological discrepancy is due to the oversegmentation caused by the watershed, the final segmentation should be a mapping of all $v_i \in V_i$ such that input vertices related to image regions corresponding to the same model object are assigned to the same model vertex. This is equivalent to merging regions of an oversegmented object into a single region.
The mapping of vertices from $G_i$ to those in $G_m$ characterizes a graph matching problem~\cite{Bunke00, ConteFSV04}. Although many mappings are possible, a desirable solution should correspond to an image partition as similar to that defined by the model as possible. Thus, to ensure that the final mapping follows the model ARG topology, the \textit{deformation} ARG $G_{d}$ is introduced. This graph is initialized as a copy of $G_m$ and it is  used to help evaluate the local deformation effect that a given assignment between a vertex $v_i \in V_i$ and another $v_m \in V_m$ induces in the model. The pursued solution is one that minimizes such effects. In the next section, we discuss how these deformations are computed and how they fit in the graph matching problem solution.

\section{The graph-matching algorithm for model-based image segmentation}
\label{sec:theory}

A segmentation of the input image according to the model under the graph-based representation is a solution for the graph matching problem between $G_i$ and $G_m$, characterized as a mapping $f: V_i \rightarrow V_m$. This implies finding a corresponding model vertex to each input vertex. Clearly, there are $|V_m|$ possible assignments for each input vertex and the decision of which to choose depends on an optimization procedure.

Let $G_d$ be an ARG initially equal to $G_m$, i.e., $V_d = V_m$, $E_d = E_m$, $\mu_d = \mu_m$, and $\nu_d = \nu_m$. Let also $v_d$ and $v_m$ be two corresponding vertices in $G_d$ and $G_m$ respectively. Suppose that an assignment from a vertex $v_i \in V_i$ to a vertex $v_m$ is under consideration. The quality of such an assignment may be assessed by computing the deformation which occurs in the model when this new vertex is merged with $v_d$, causing the attributes of $\mu_d(v_d)$ and $\nu_d(v_d)$ to be fused with those from $v_i$. After such merge, $G_d$ becomes a distorted version of the model in which:

\begin{equation}
  \label{eq:newmu}
  \mu_d(v_d) = (\frac{(R_{v_d} + R_{v_i})}{2}, \frac{(G_{v_m} + G_{v_i})}{2}, \frac{(B_{v_m} + B_{v_i})}{2})
\end{equation} 

\noindent and 

\begin{equation}
  \label{eq:newnu}
  \nu_d(e) = (p_{w_d} - p_{v_d}) / (2d_{max})
\end{equation}

\noindent $\forall e \in E_d(v_d) = \{e \in E_d: e = (v_d, w_d)$ or $e = (w_d, v_d), w_d \in V_d\}$ and with $p_{v_d} = \frac{p_{v_m} + p_{v_i}}{2}$. 

The impact of such deformation is then measured according to the following \textit{cost function}:
\begin{equation}
\label{func:objective1}
f(G_d, G_m) = \alpha c_V(v_d, v_m) + \frac{(1-\alpha)}{|E_d(v_d)|} \sum_{e \in E_{d}(v_d)} c_E(e, e_m)
\end{equation}

The term $c_V(v_d, v_m)$ is a measure of the deformation between the object attributes of $v_d$ and $v_m$ and it is defined as:

\begin{equation}
  \label{eq:objcost}
  c_V(v_d, v_m) = \sqrt{\sum_{C = R,G,B} (C_{v_d} - C_{v_m})^2}
\end{equation}

Although the RGB color space was chosen to describe the color appearance feature of an object, other color spaces, such as the Lab, might as well be used with appropriate metrics adaptation.



Similarly, if $e \in E_d(v(d))$ and $e_m \in E_m$ is its corresponding edge, then $c_E(e, e_m)$ is a measure of the deformation between both edges defined as:

\begin{equation}
  \label{eq:edgecost}
  c_E(e, e_m) = \gamma_E \frac{ | \cos(\theta) - 1 | }{2} + (1-\gamma_E) | \|\nu(e)\| - \|\nu(e_m)\| |
\end{equation}

 The value $\theta$ is the angle between $\nu(e)$ and $\nu(e_m)$, whereas, the parameter $\gamma_E$, $0 \leq \gamma_E \leq 1$, controls the weights of the modulus and angular dissimilarities. Thus, the total impact caused on the edges directly connected by $v_d$ is computed as the modulus and angular differences between the relational attribute vectors of each pair $(e, e_m)$.

Therefore, the cost function measures how the merging of a vertex $v_i$ with a copy $v_d$ of a model vertex affects the local structure of the graph, as well as the appearance attributes it holds. The parameter $\alpha$, $0 \leq \alpha \leq 1$, controls the importance of the appearance or structural effects of all vertex mappings.    





\section{The optimization algorithm}
\label{sec:algo}

To map $G_i$ to $G_m$ and estimate the adequacy of each vertex assignment using $G_d$, the following algorithm was devised:\\

\hrule 
\begin{enumerate}
\item define $\alpha$
\item define $\gamma$
\item for each vertex $v_i \in V_i$
\item \ \ \ \ \ \ $f_{min} \leftarrow \infty$
\item \ \ \ \ \ \ $minlbl \leftarrow -1$
\item \ \ \ \ \ \ for each vertex $v_d \in V_{d}$ 
\item \ \ \ \ \ \ $G_d \leftarrow G_m$
\item \ \ \ \ \ \ $\mu(v_d) (\frac{R_{v_d} + R_{v_m}}{2}, \frac{G_{v_d} + G_{v_m}}{2}, \frac{B_{v_d} + B_{v_m}}{2})$ 
\item \ \ \ \ \ for each $e \in E_d((v_d)$ s.t. $e = (v_d, w)$ or $e = (w, v_d)$, $w \in V_d$
\item \ \ \ \ \ \ \ \ \ \ $\nu(e) \leftarrow$ vector between centroids of $v_d$ and $w$
\item \ \ \ \ compute the value $f$ of the cost function (eq.~\ref{func:objective1}) $f(G_{d}, G_{m})$
\item \ \ \ \ \ \ if $f < f_{min}$ 
\item \ \ \ \ \ \ \ \ \ \ $f_{min} \leftarrow f$
\item \ \ \ \ \ \ \ \ \ \ $minlbl \leftarrow v_d$
\item \ \ \ \ \ \ label($v_i$) $\leftarrow minlbl$
\end{enumerate}
\hrule \ \\

All possible assignments for each vertex $v_i \in V_i$ are evaluated and the final label of $v_i$ corresponds to the model vertex which was affected the least by the deformation caused by the introduction of $v_i$. At each iteration, a vertex from $V_i$ receives a label. $f_{min}$ is the minimum deformation cost obtained so far and $minlbl$ is the corresponding model vertex to which $v_i$ was mapped resulting in such minimal deformation. The output of the algorithm is a mapping of all vertices of $V_i$ to vertices of $V_m$.

\section{Experimental results}
\label{sec:results}

\begin{figure}[t]
\begin{center}
\begin{tabular}{cc}
   \includegraphics[width=2.7cm]{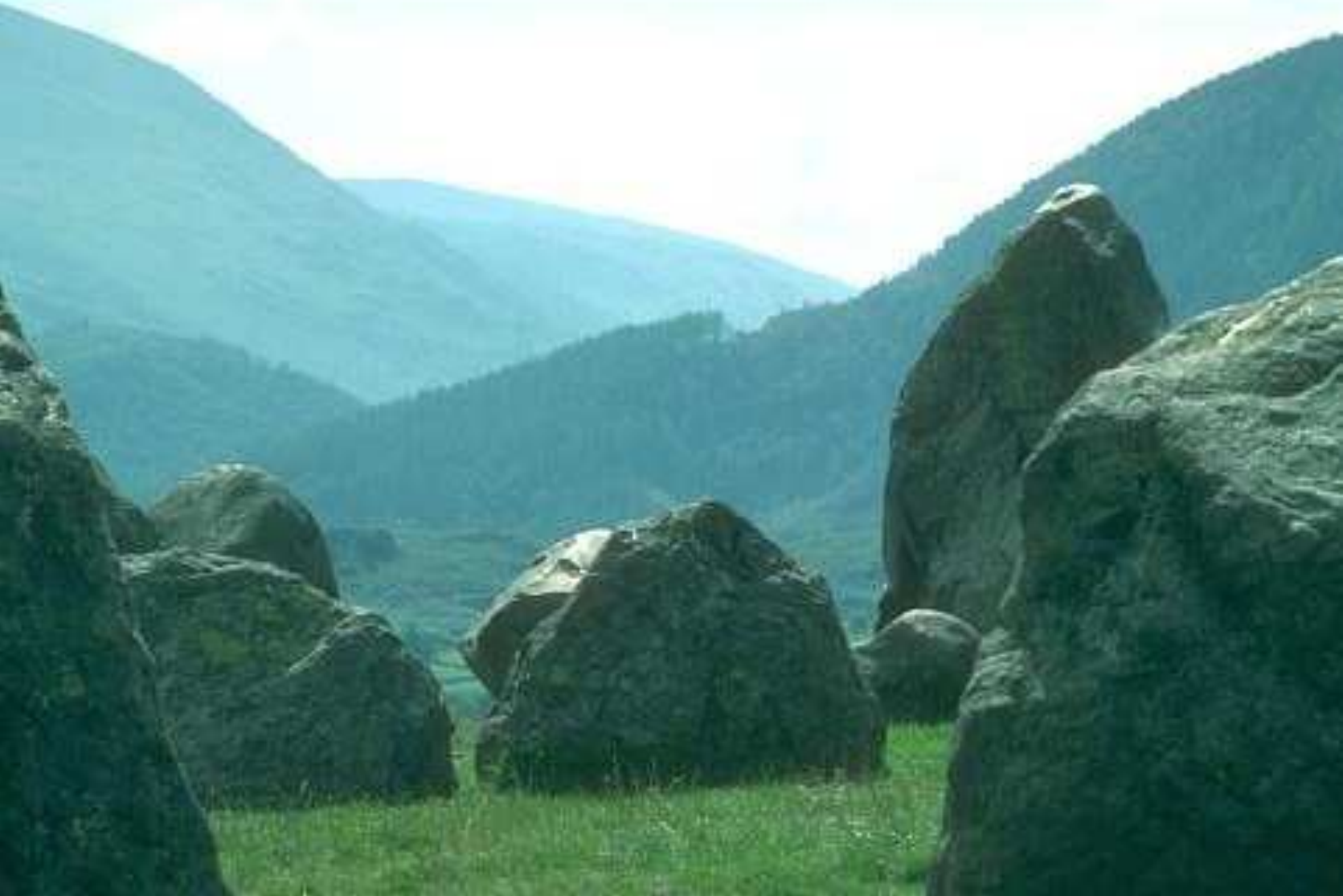}
   \includegraphics[width=2.7cm]{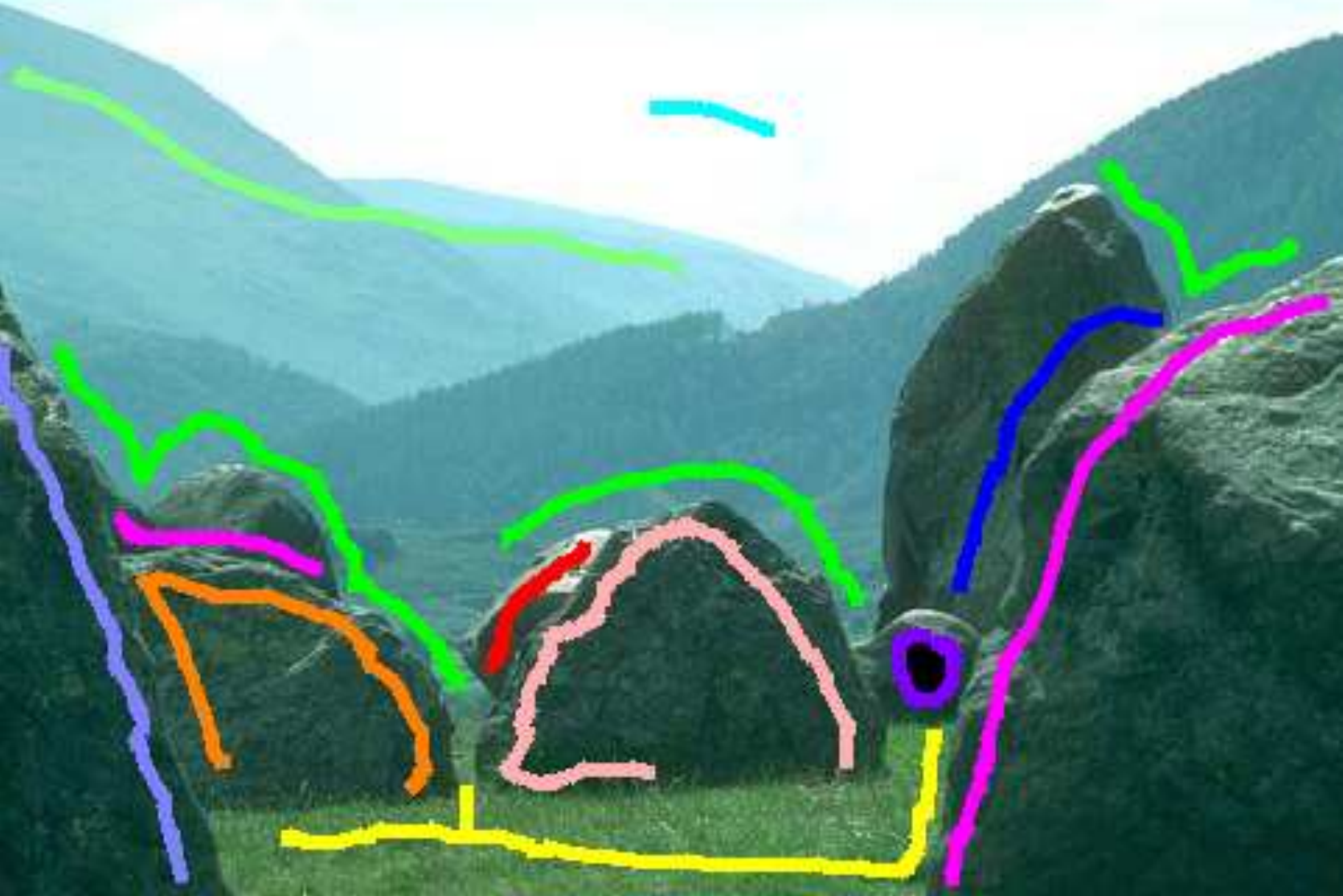} 
   \includegraphics[width=2.7cm]{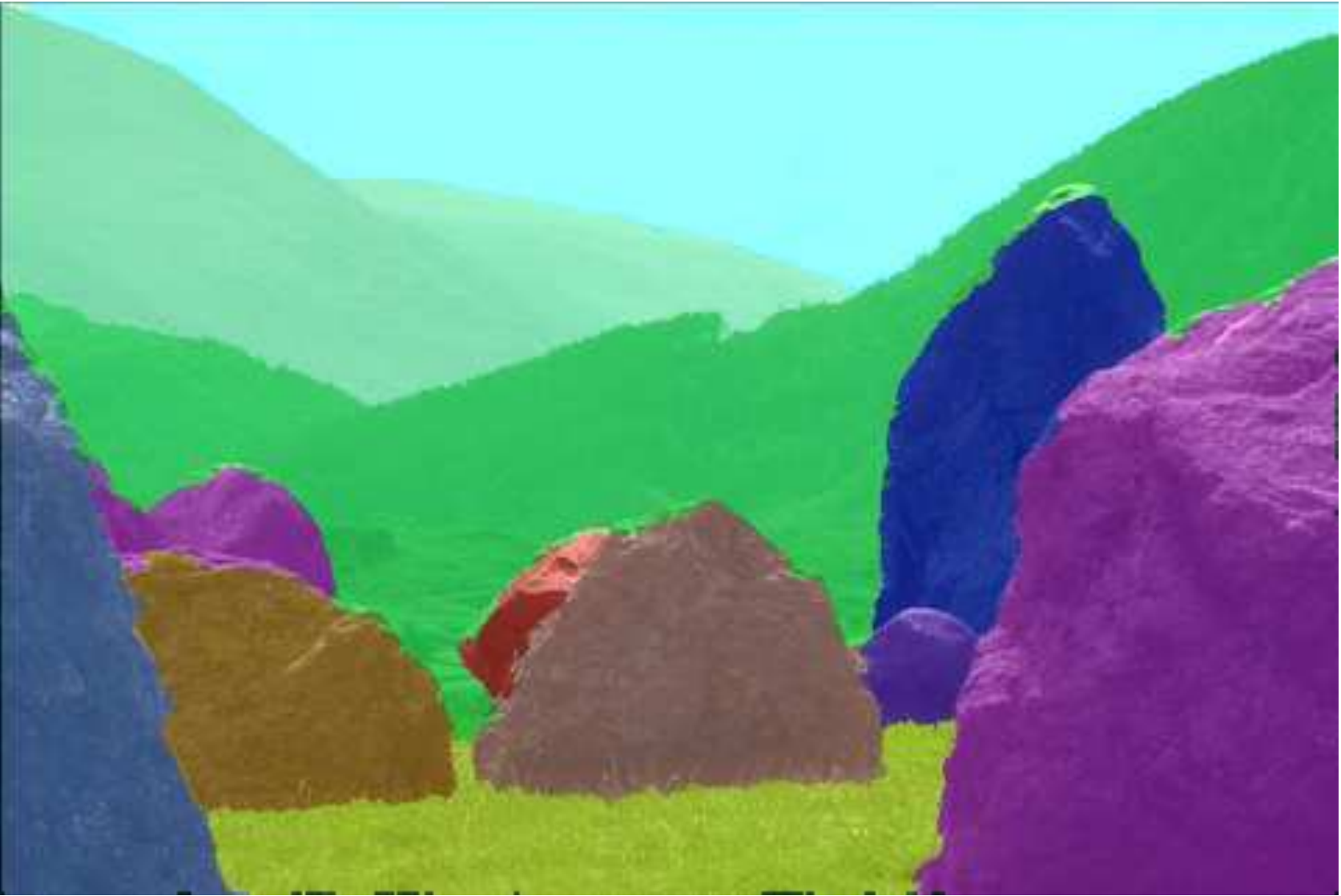}\\
   \includegraphics[width=2.7cm]{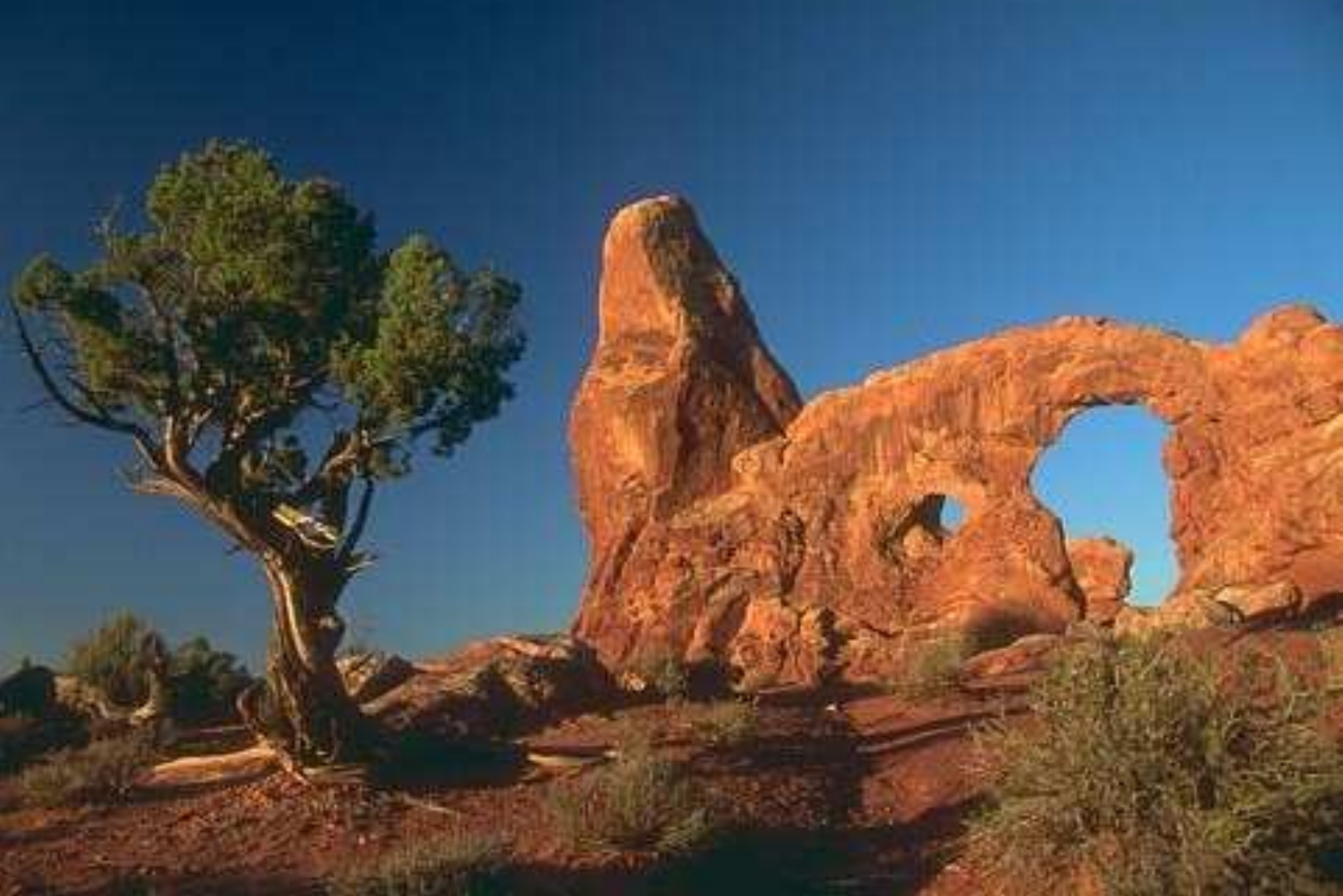}
   \includegraphics[width=2.7cm]{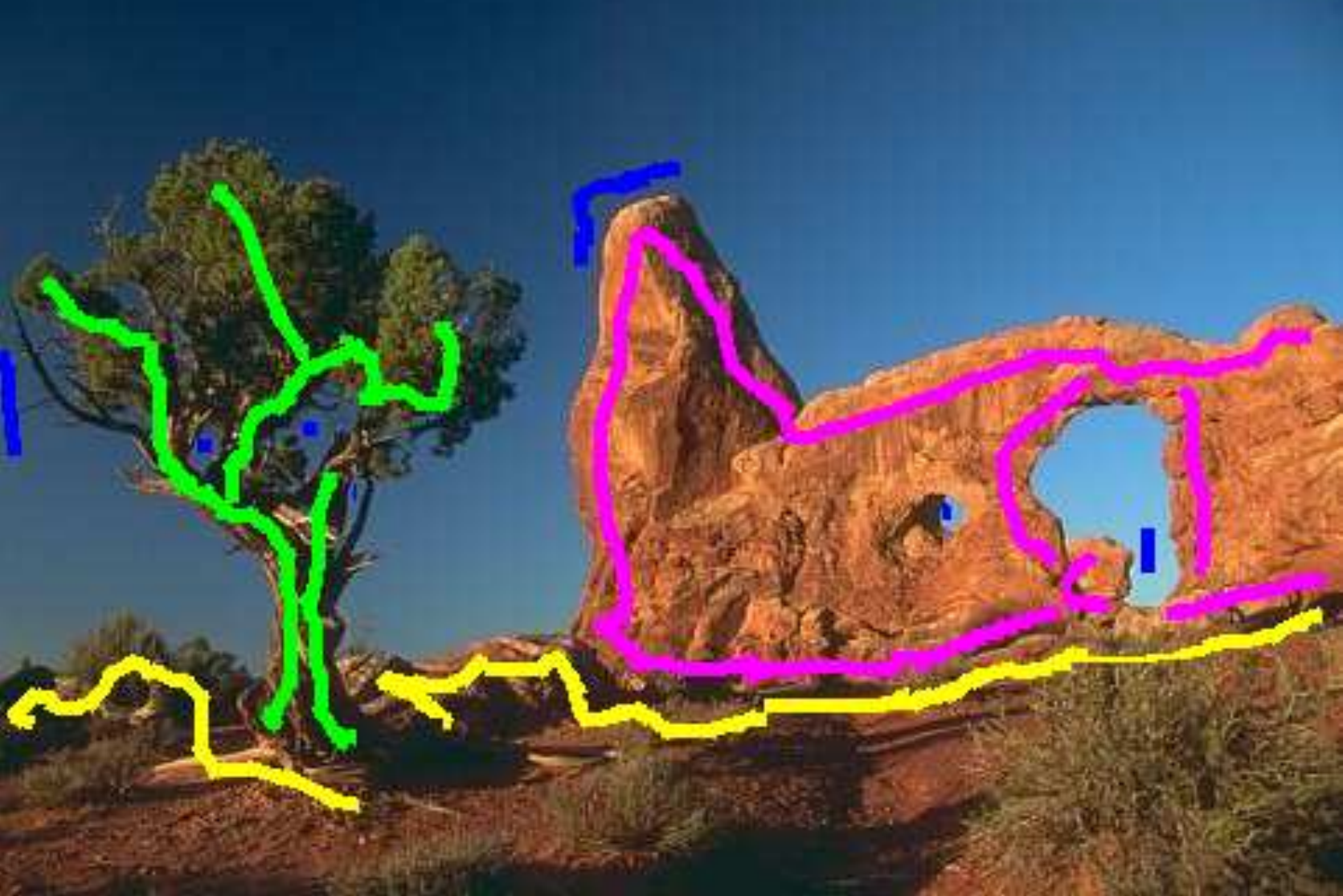} 
   \includegraphics[width=2.7cm]{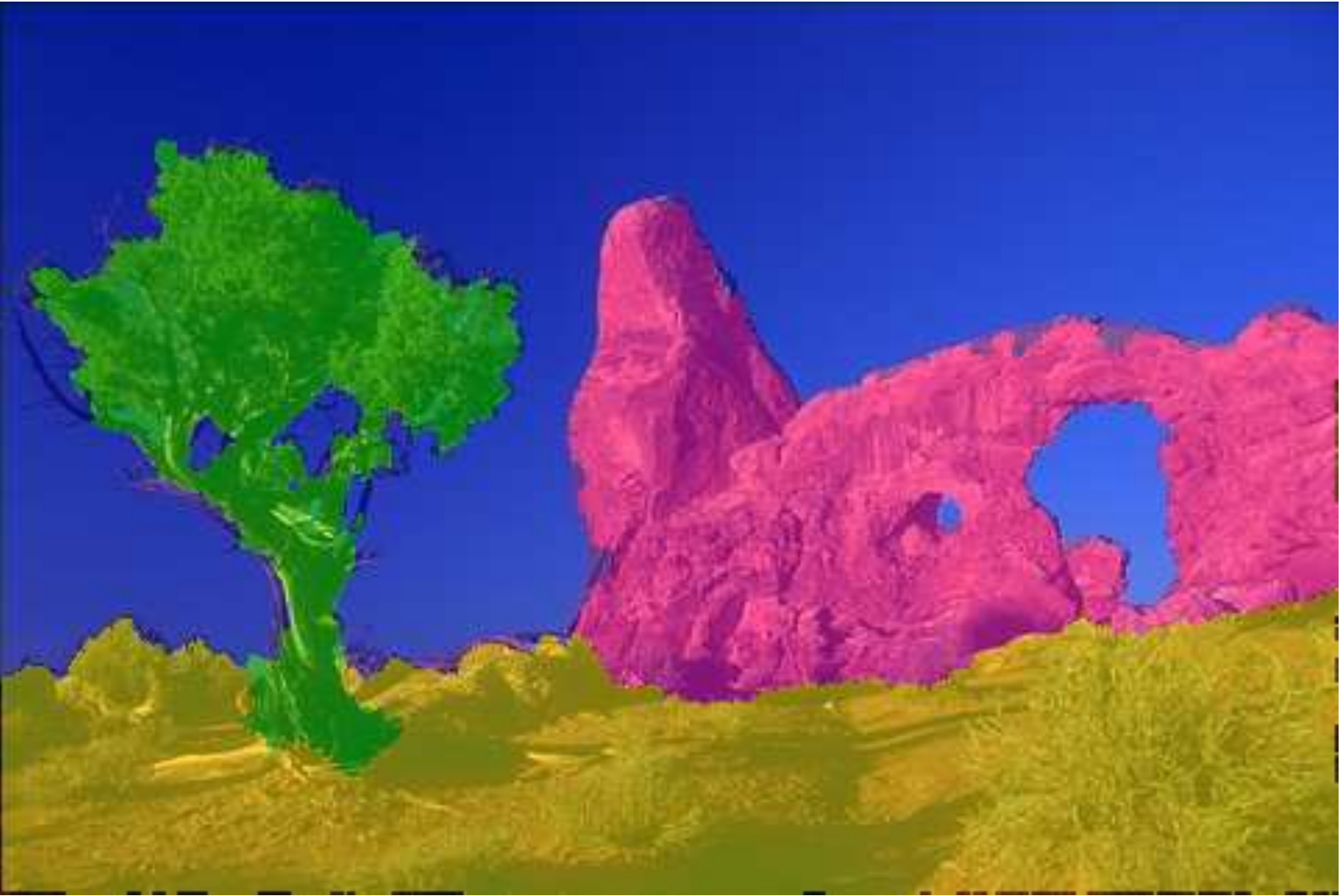}\\
   \includegraphics[width=2.7cm]{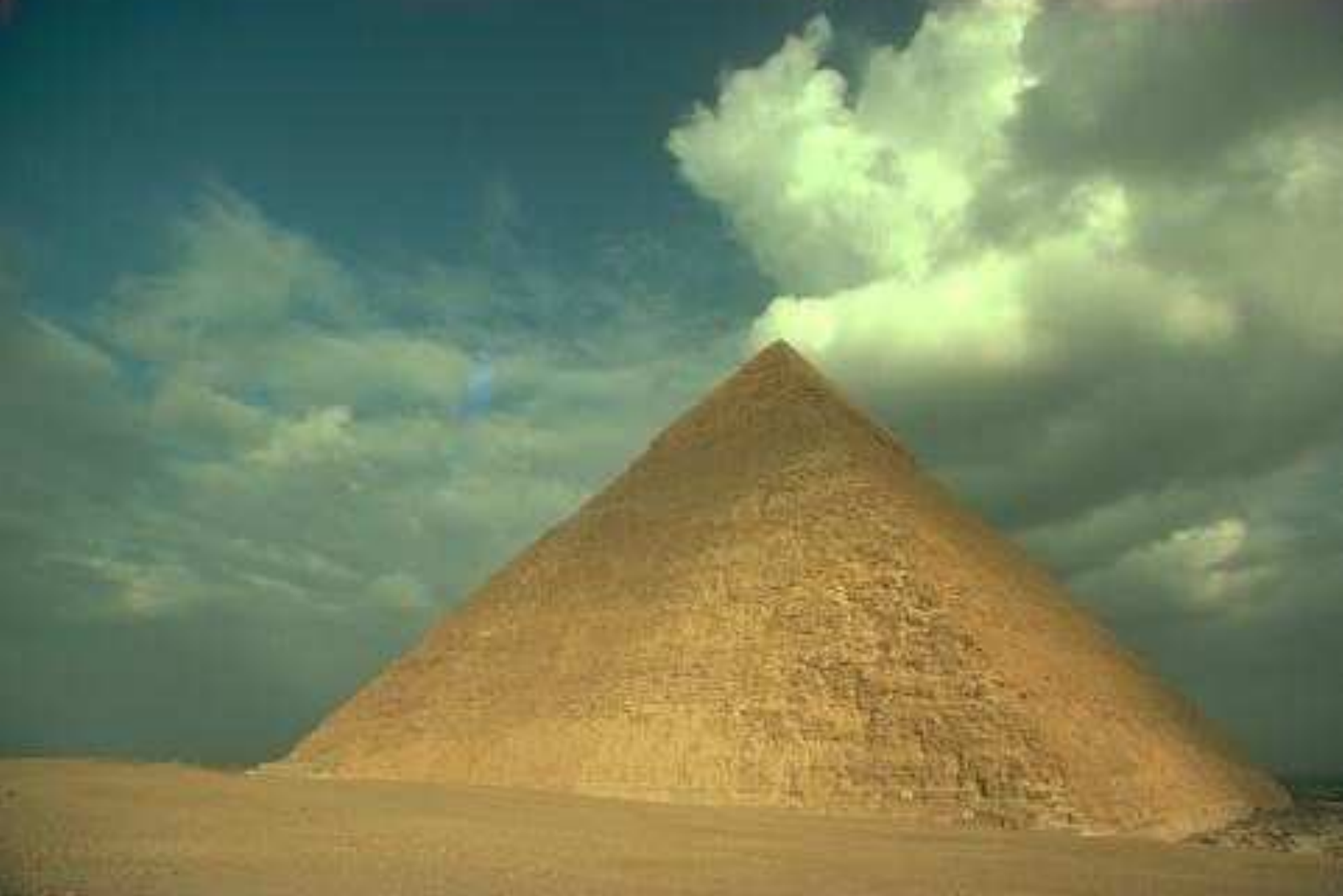}
   \includegraphics[width=2.7cm]{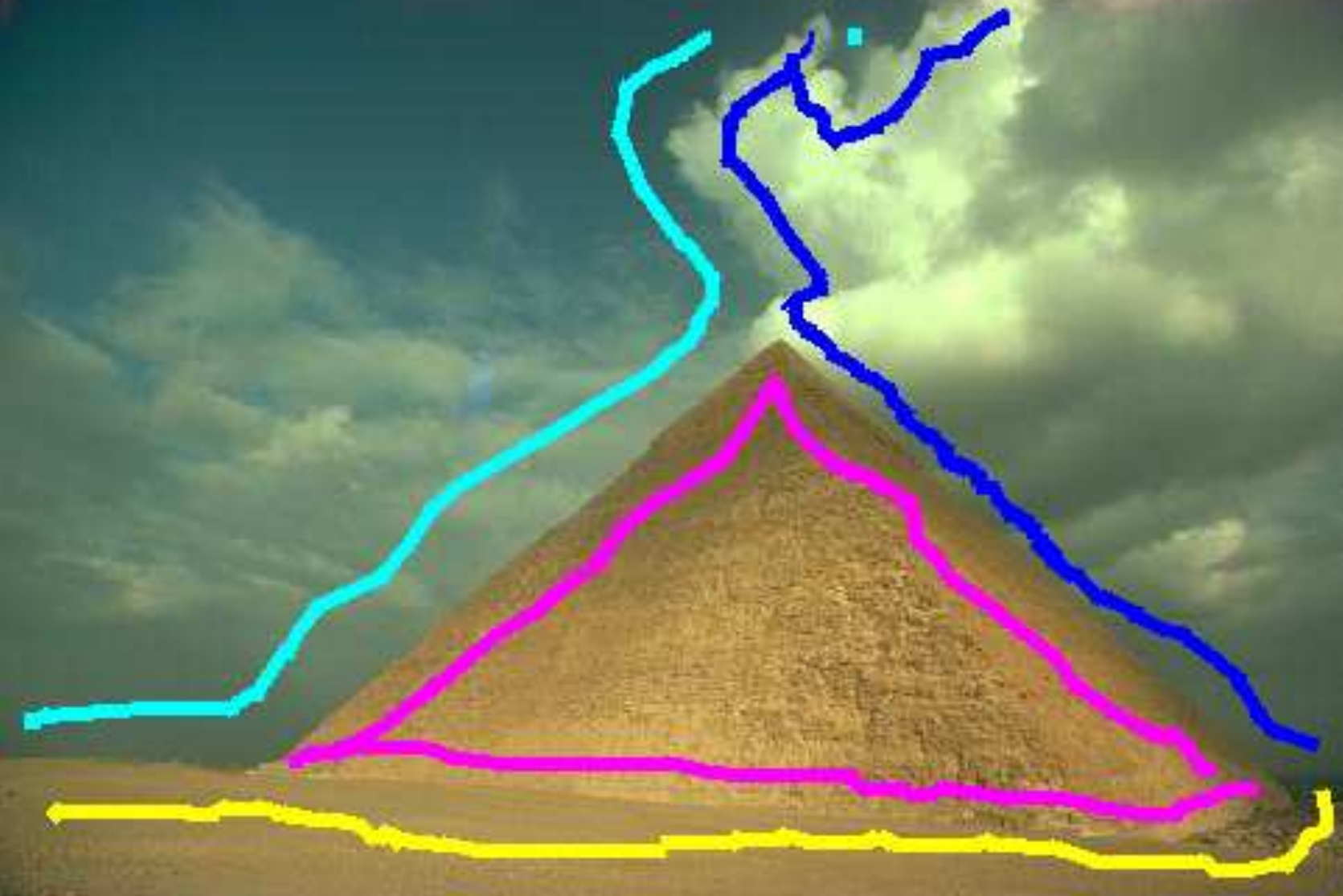} 
   \includegraphics[width=2.7cm]{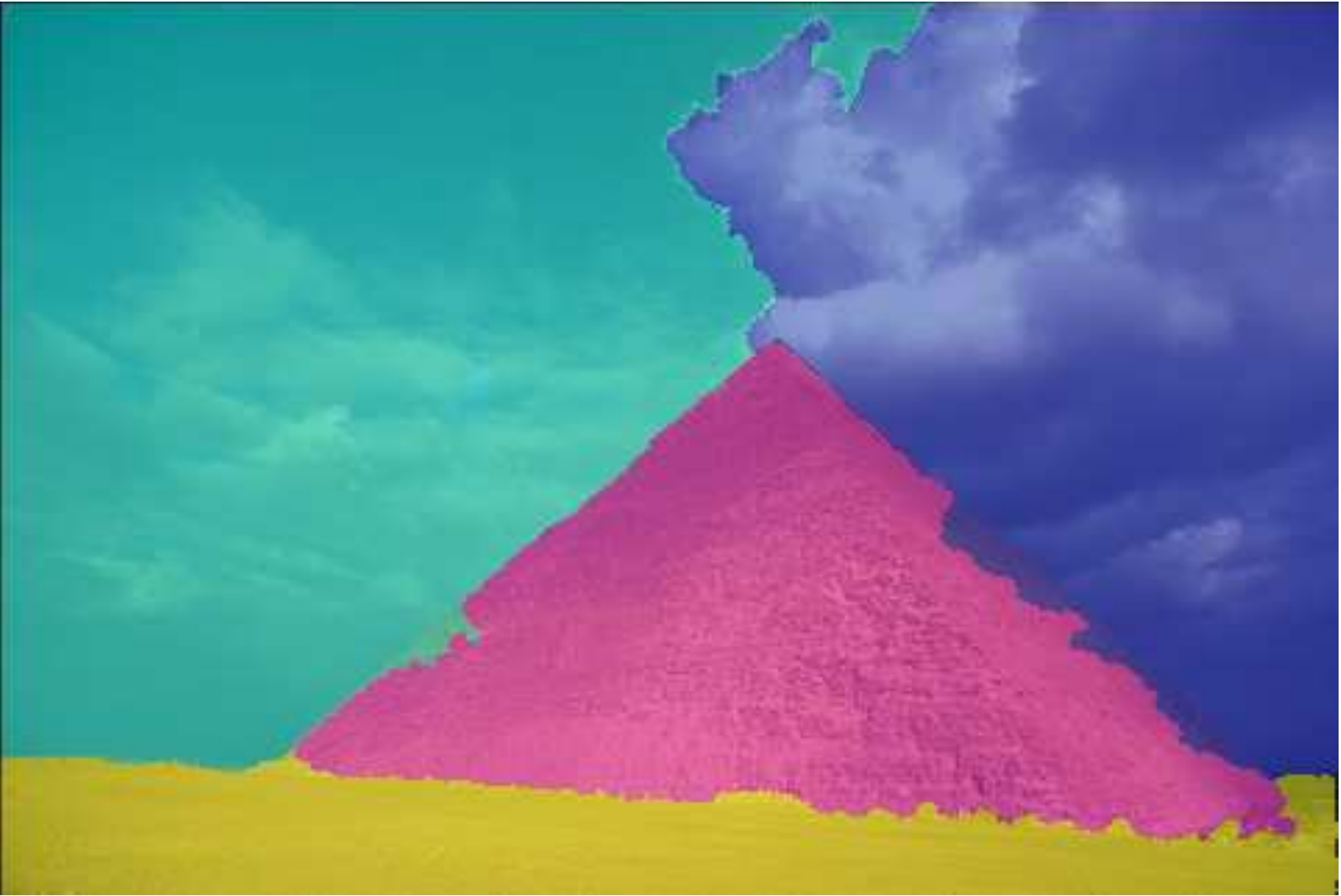}\\
   \includegraphics[height=3.2cm]{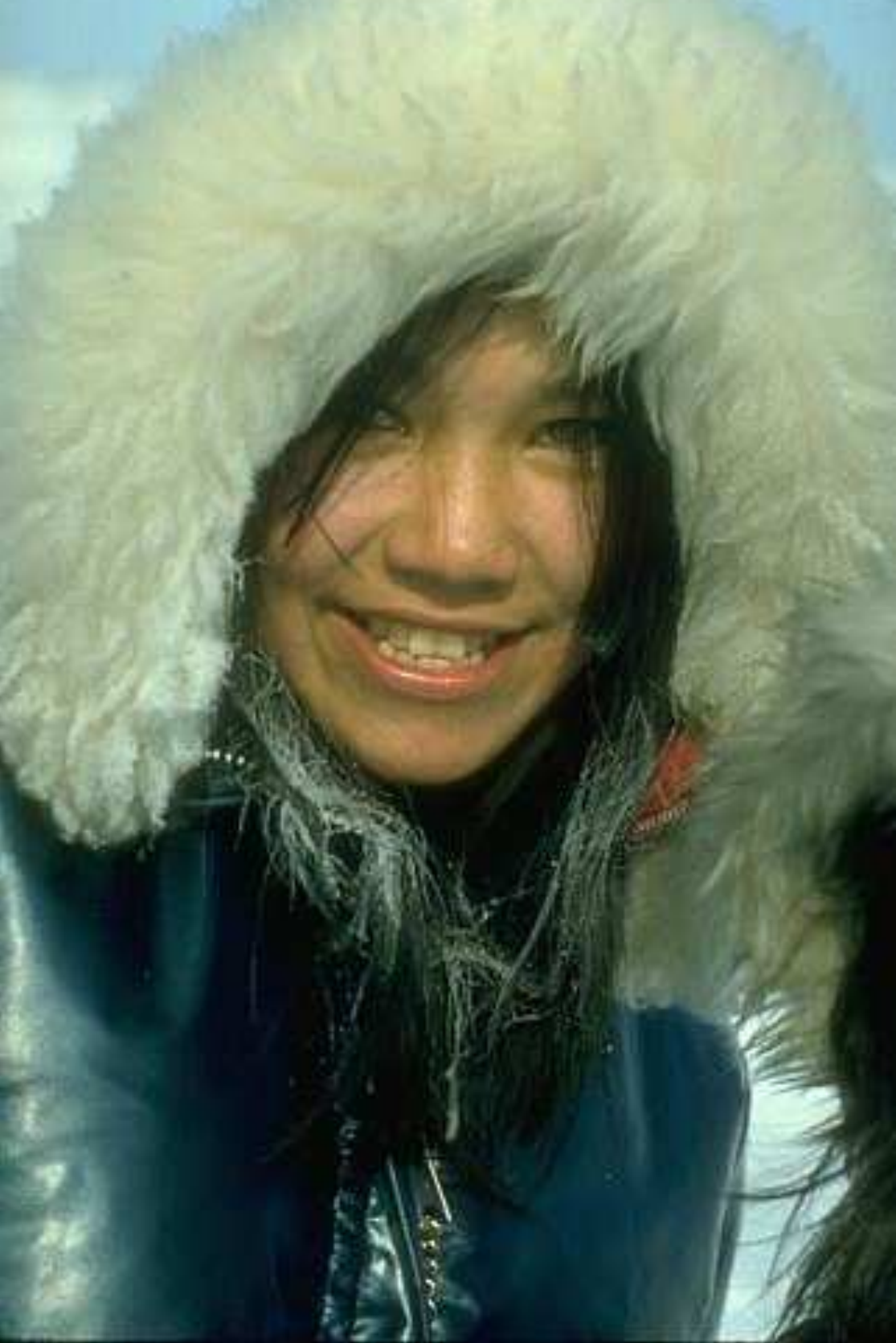}
   \includegraphics[height=3.2cm]{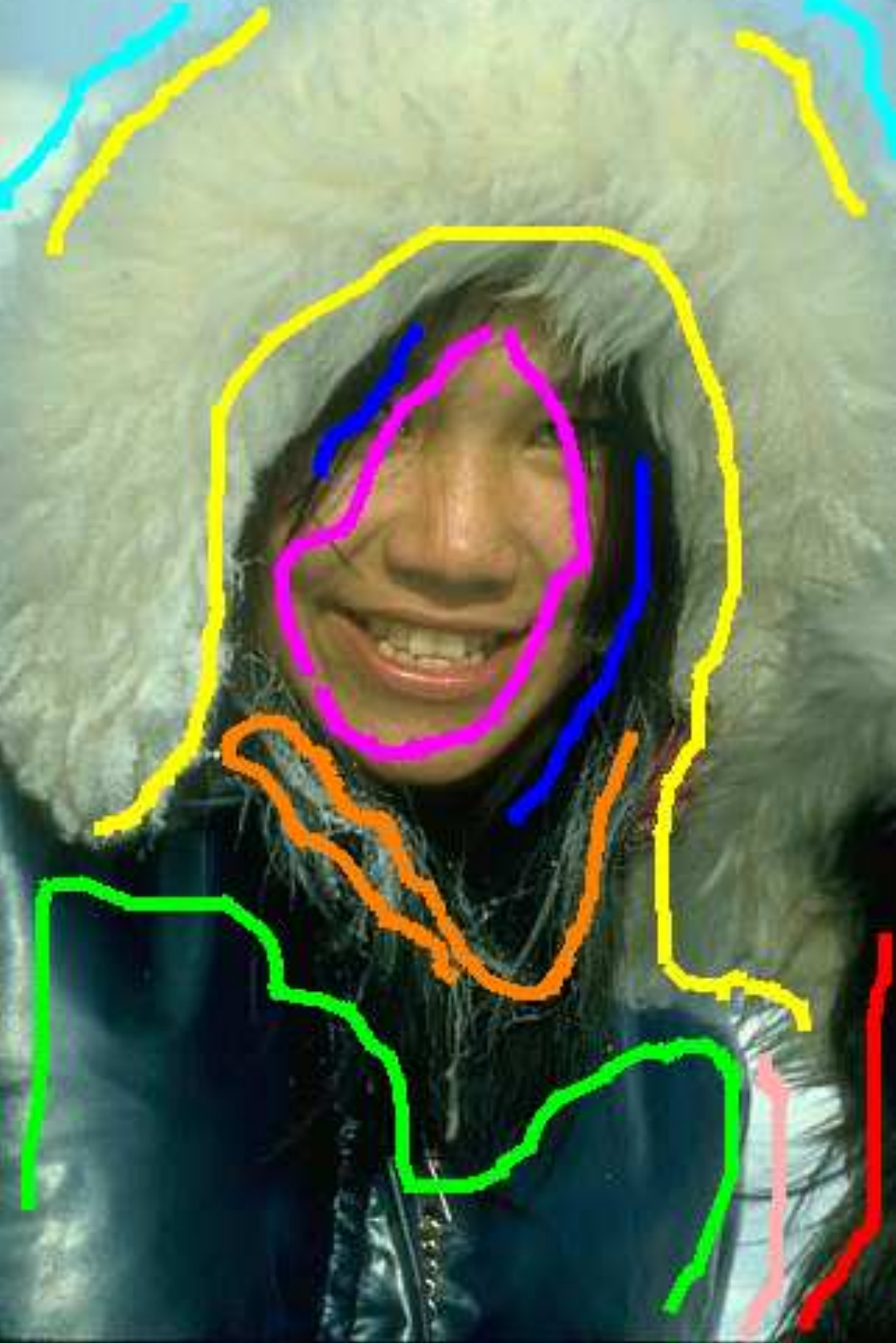} 
   \includegraphics[height=3.2cm]{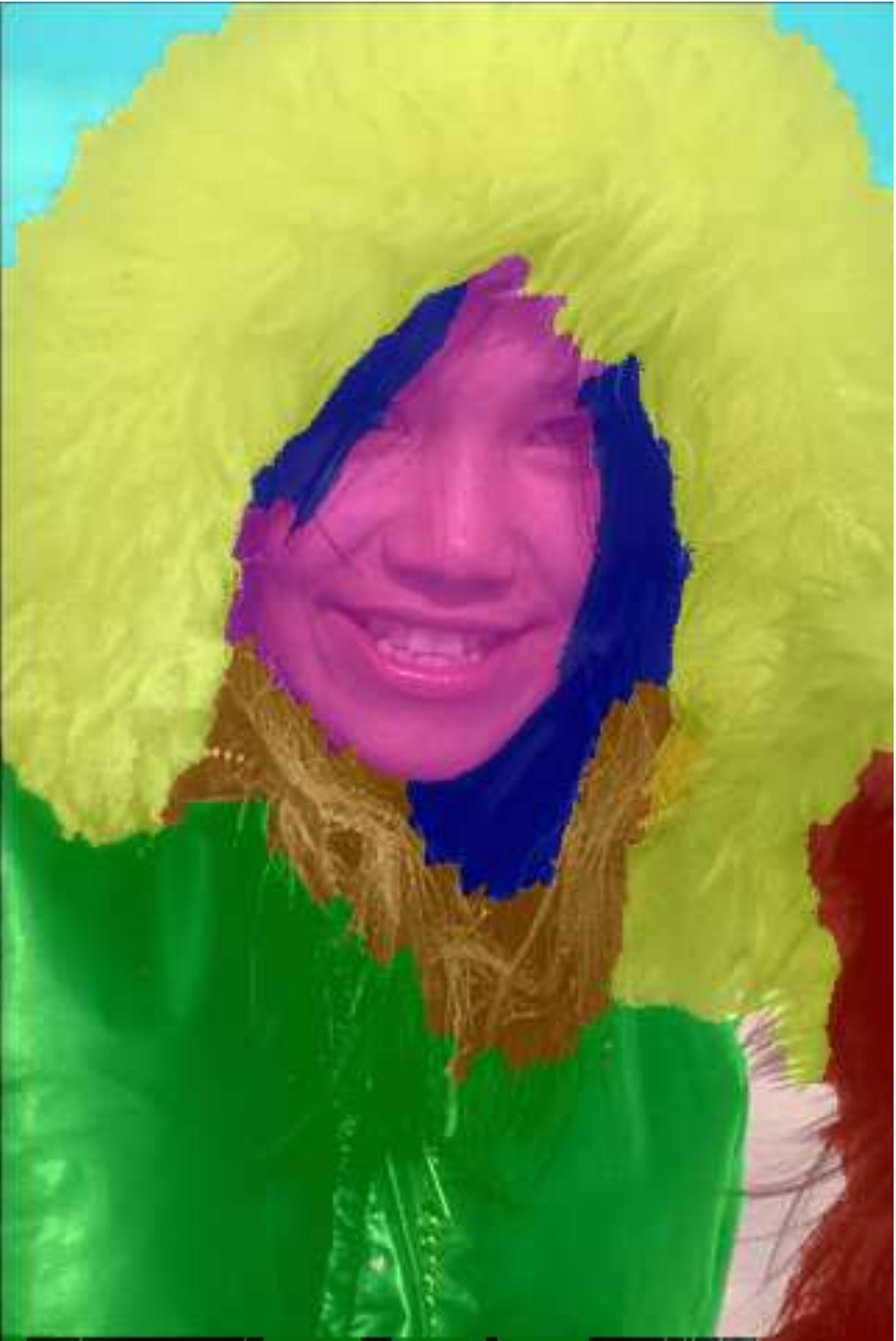}\\
\end{tabular}
\end{center}

\caption{\label{fig:results} Sample segmentation results: original images (left col.), images with user-input strokes defining regions of interest (center col.) and corresponding resulting partitions (right col.).}
\end{figure}

In order to test the presented technique, a Java application was designed and implemented~\footnote{\tiny{Please refer to the accompanying demo video.}}. Its interface allows the user to load images to be segmented, define a model according to traces drawn over different regions of interest, and choose the $\alpha$ and $\gamma$ parameters of the cost function (eq.~\ref{func:objective1}), therefore specifying how to favor structure against appearance features. 

Figure~\ref{fig:results} shows a few segmentation results for the application of the methodology to natural color images obtained from the Berkeley Image Segmentation Database~\footnote{\tiny{\texttt{http://www.eecs.berkeley.edu/Research/Projects/CS/vision/grouping/segbench/}}}. All resulting images depict the final regions labelled according to the color of the traces defined by the user. Transparency was used over the labelled segmented images in order to visualize more precisely what areas have been classified as a given object. Although certain image regions, such as the mountains and the face, present high variability due to textures or different objects grouped as one in the model, the final segmentation remains accurate and robust thanks to the structural constraints embedded in the model.

\begin{figure}[t]
\begin{center}
\begin{tabular}{l}
   \includegraphics[width=8cm]{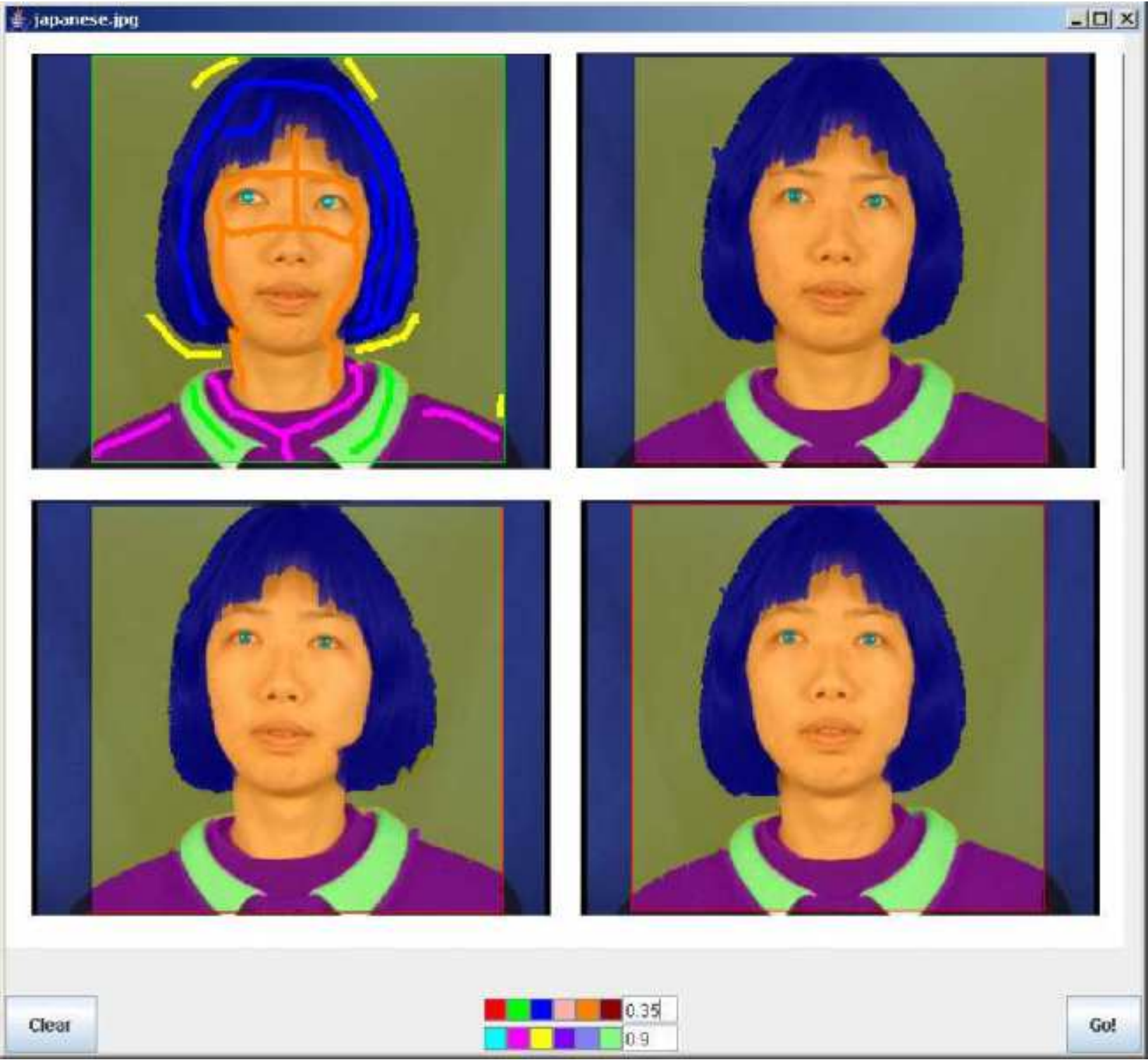} \\
   \includegraphics[width=8cm]{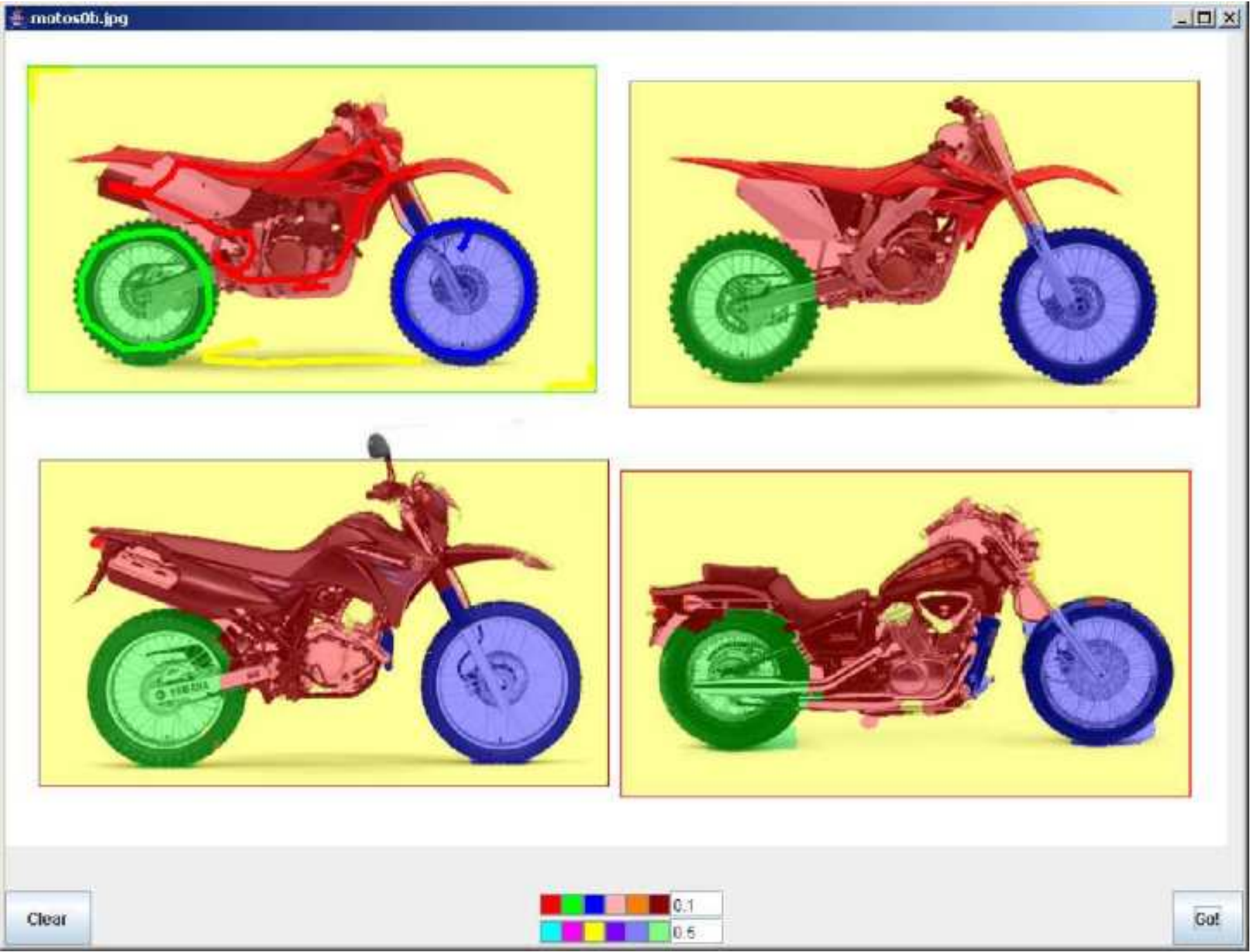} 
\end{tabular}
\end{center}

\caption{\label{fig:results2} Sample segmentation results after applying the same user-defined stroke model (top left image in each set) to different images.}
\end{figure}

Tests for the reusability and robustness of the model were also performed on sample frames of a moving head video (fig.~\ref{fig:results2} top) from the XM2VTS Database \footnote{\tiny{\texttt{http://www.ee.surrey.ac.uk/CVSSP/xm2vtsdb/}}} and on a set of similar images retrieved randomly from the web (fig.~\ref{fig:results2} bottom). Each model was defined once by the user in the first image of each set and then applied to the other similar images. The latter step is interactively accomplished as follows: once the user draws the traces over the first image, a minimum enclosing rectangle of the strokes is automatically defined. This rectangle, called a \textit{stamp}, can later be applied by the user to other images and segmentation can be performed within such area. 

\begin{figure}[t]
\begin{center}
   \includegraphics[width=8cm]{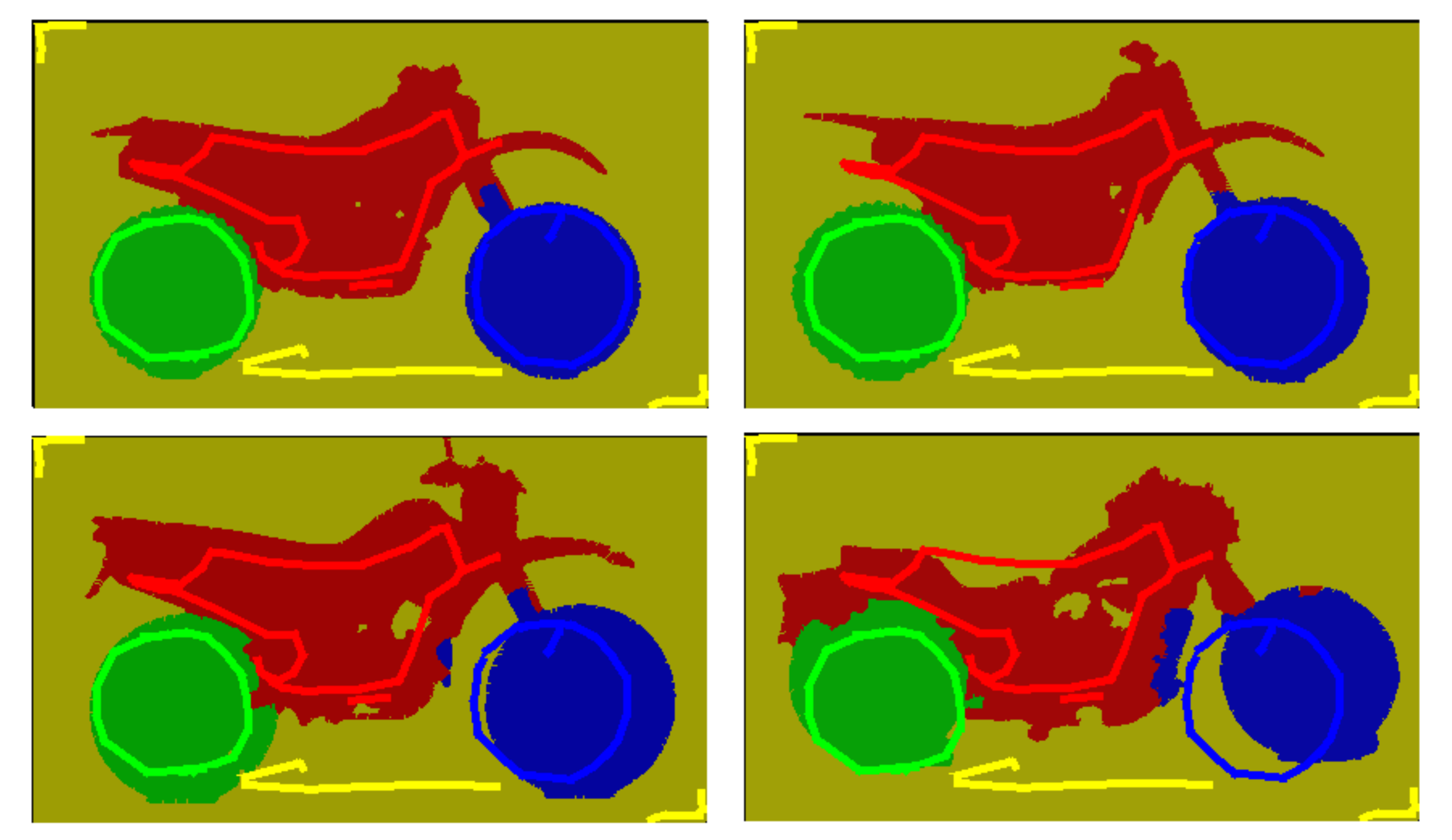}
\end{center}

\caption{\label{fig:results3} Replication of the model: simply reusing the user-defined strokes over different images does not guarantee that the model to be derived is always consistent, since the strokes could fall over distinct objects from one image to another and the final segmentation would be compromised.}
\end{figure}

Note that the model ARG is derived only once for the first image and then used in the segmentation process of all other input images. It is important to notice that simply applying the same strokes to the other images would not produce the same model as the one obtained for the first time, as fig.~\ref{fig:results3} depicts. This also shows that the model is robust enough to treat small variances presented by the input images under analysis. For each image set, the overall structure remained similar and the segmentation was once again satisfactory even though the model was derived from an image with different appearance features.

The new algorithm proposed for the graph matching step presents faster performance than that of the algorithm reported in~\cite{consulcesarblochICIP07}. While the present algorithm runs in time proportional to $O(|V_i||V_m|)$, the other is bounded by a function $\Omega(|V_i||V_m|^2)$. Besides this, the optimization algorithm does not depend on the order in which vertices from the input are labelled, since each vertex is treated separately when analysed during the graph matching step.


\section{Conclusion}
\label{sec:conclusion}

This paper proposed a novel algorithm for performing interactive model-based image segmentation using attributed relational graphs to represent both model and input images. This approach allows the usage of information ranging from appearance features to structural constraints. Topological differences between graphs are dealt with by means of a deformation ARG, a structure which allowed the design of an optimization algorithm for graph matching that evaluates possible solutions according to local impacts (or deformations) they determine on the model. The faster performance of the algorithm in comparison with the one proposed in~\cite{consulcesarblochICIP07}, the reusability of the model graph when segmenting several images, as well as the satisfying quality of the results due to the adequate use of structural information, characterize the main contributions of the method. 

Our ongoing work is devoted to reducing interaction when reusing the model to segment various images. For now, it is required that the user places the stamp over the area of interest of the image. In the future, we hope to be able to apply the model ARG without the need of this interactive positional information. This shall be accomplished through the investigation of MAP-MRF methods applied within this framework in order to make more robust models and improve segmentation quality under different conditions such as object translation and rotation. Furthermore, we intend to perform a quantitative study to compare the accuracy of our results with those of other related methods.


{\small
\bibliographystyle{ieee}
\bibliography{cvpr08}
}



\end{document}